\pdfoutput=1

\documentclass[11pt]{article}

\usepackage{authblk}

\usepackage{naacl2021}

\usepackage{times}
\usepackage{latexsym}

\usepackage[T1]{fontenc}

\usepackage[utf8]{inputenc}

\usepackage{microtype}

\usepackage{booktabs}
\usepackage{amsmath}
\usepackage{enumerate}
\usepackage{cleveref}
\usepackage{multirow}
\usepackage{bbding}
\usepackage{mathtools}
\usepackage{subfig}

\newlength\lengtha 
\newlength\lengthb 
\newlength\lengthc 

\DeclarePairedDelimiterX{\infdivx}[2]{(}{)}{%
  #1\;\delimsize\|\;#2%
}

\usepackage{graphicx}
\graphicspath{figures/}

\newcommand{\robertazs}[0]{RoBERTa\textsubscript{Zero-shot}}

\newcommand{\conceptmax}[0]{\textsc{ConceptMax}}
\newcommand{\conceptinject}[0]{\textsc{ConceptInject}}

\newcommand{\pep}[0]{\textsc{Pep-3k}}

\newcommand{\ccd}[0]{CC$\Delta$}
\newcommand{\ler}[0]{LER}
\newcommand{\mlp}[0]{GloVe+MLP}

\title{Modeling Event Plausibility with Consistent Conceptual Abstraction}

\author[1]{Ian Porada}
\author[2]{Kaheer Suleman}
\author[2]{Adam Trischler}
\author[1]{Jackie Chi Kit Cheung}
\affil[1]{Mila, McGill University}
\affil[ ]{{\tt \{ian.porada@mail, jcheung@cs\}.mcgill.ca}}
\affil[2]{Microsoft Research Montr\'eal}
\affil[ ]{{\tt \{kasulema, adam.trischler\}@microsoft.com}}

\begin{document}
\maketitle
\begin{abstract}

Understanding natural language requires common sense, one aspect of which is the ability to discern the plausibility of events. While distributional models---most recently pre-trained, Transformer language models---have demonstrated improvements in modeling event plausibility, their performance still falls short of humans'. In this work, we show that Transformer-based plausibility models are markedly inconsistent across the conceptual classes of a lexical hierarchy, inferring that ``a person breathing'' is plausible while ``a dentist breathing'' is not, for example. We find this inconsistency persists even when models are softly injected with lexical knowledge, and we present a simple post-hoc method of forcing model consistency that improves correlation with human plausibility judgements.
\end{abstract}

\section{Introduction}

Of the following events, a human reader can easily discern that (\ref{itm:typical}) and (\ref{itm:atypical}) are semantically plausible, while (\ref{itm:implaus}) is nonsensical.

\begin{enumerate}[(1)]
\setlength\itemsep{0.2em}
    \item \label{itm:typical} The person breathes the air.
    \item \label{itm:atypical} The dentist breathes the helium.
    \item \label{itm:implaus} The thought breathes the car.
\end{enumerate}

This ability is required for understanding natural language: specifically, modeling \textit{selectional preference}---the semantic plausibility of predicate-argument structures---is known to be implicit in discriminative tasks such as coreference resolution \cite{HOBBS1978311, ido-anaphora-res, zhang-etal-2019-knowledge}, word sense disambiguation \cite{resnik-1997-disambig, mccarthy-disambig}, textual entailment \cite{zanzotto-etal-2006-discovering, pantel-etal-2007-isp}, and semantic role labeling \cite{gildea-jurafsky-2002-automatic, zapirain-etal-2013-selectional}.
 
More broadly, modeling semantic plausibility is a necessary component of generative inferences such as conditional commonsense inference \cite{Gordon2011SemEval2012T7, zhang-etal-2017-ordinal}, abductive commonsense reasoning \cite{Bhagavatula2020Abductive}, and commonsense knowledge acquisition \cite{ijcai2020-554, hwang2020cometatomic}.

\begin{figure}
\centering
  \includegraphics[width=7.25cm]{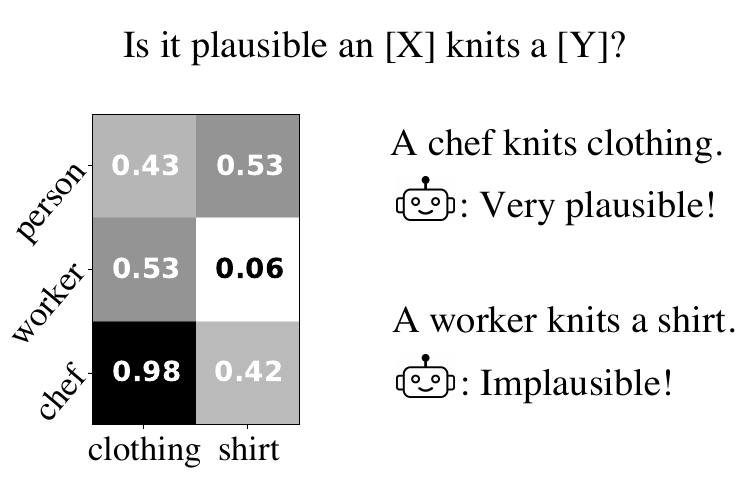}
  \caption{
  Elements in the matrix are the relative plausibility score for the event ``an [X] knits a [Y]'' as output by a RoBERTa model fine-tuned to model plausibility. [X] and [Y] correspond to the label of the row and column, respectively. Model scores are inconsistent with respect to the two events shown on the right.
  }
  \label{fig:intro-example}
\end{figure}

Learning to model semantic plausibility is a difficult problem for several reasons. First, language is sparse, so most events will not be attested even in a large corpus. Second, plausibility relates to likelihood in the world, which is distinct from the likelihood of an event occurring in language. Third, plausibility reflects human intuition, and thus modeling plausibility at its extreme requires ``the entire representational arsenal that people use in understanding language, ranging from social mores to naive physics'' \cite{resnik_1996}.

A key property of plausibility is that the plausibility of an event is generally consistent across some appropriate level of abstraction. For example, events of the conceptual form ``the [\textsc{person}] breathes the [\textsc{gas}]'' are consistently plausible. Plausibility judgments follow this pattern because people understand that similar concept classes share similar affordances.

Furthermore, the change in plausibility between levels of abstraction is often consistent. Consider that as we abstract from ``person breathes'' to ``organism breathes'' to ``entity breathes,'' plausibility consistently decreases.

In this paper, we investigate whether state-of-the-art plausibility models based on fine-tuning Transformer language models likewise exhibit these types of consistency. As we will show, inconsistency is a significant issue in existing models which results in erroneous predictions (See Figure~\ref{fig:intro-example} for an example).

To address this issue, we explore two methods that endow Transformer-based plausibility models with knowledge of a lexical hierarchy---our hypothesis being that these methods might correct conceptual inconsistency without over-generalizing. The first method makes no a priori assumptions as to how the model should generalize and simply provides lexical knowledge as an additional input to the model. The second explicitly enforces conceptual consistency across a lexical hierarchy by taking the plausibility of an event to be a maximum over the plausibility of all conceptual abstractions of the event.

We find that only the second proposed method sufficiently biases the model to more accurately correlate with human plausibility judgments. This finding encourages future work that forces Transformer models to make more discrete abstractions in order to better model plausibility.

We focus our analysis on simple events in English represented as subject-verb-object (s-v-o) triples, and we evaluate models by correlation with two datasets of human plausibility judgements. Our models build off of RoBERTa \cite{Liu2019RoBERTaAR}, a pre-trained Transformer masked language model.\footnote{Our implementation and data will be available at \url{https://github.com/ianporada/modeling_event_plausibility}} We use WordNet 3.1 \cite{wordnet} hypernymy relations as a lexical hierarchy.

Concretely, our contributions are:
\begin{itemize}
    \setlength\itemsep{0.0em}
    \item We evaluate the state of the art in modeling plausibility, both in terms of correlation with human judgements and consistency across a lexical hierarchy.
    \item We propose two measures of the consistency of plausibility estimates across conceptual abstractions.
    \item We show that injecting lexical knowledge into a plausibility model does not overcome conceptual inconsistency.
    \item We present a post-hoc method of generalizing plausibility estimates over a lexical hierarchy that is necessarily consistent and improves correlation with human plausibility judgements.
\end{itemize}

\section{Related Work}

\begin{figure}[h]
\centering
  \includegraphics[width=5cm]{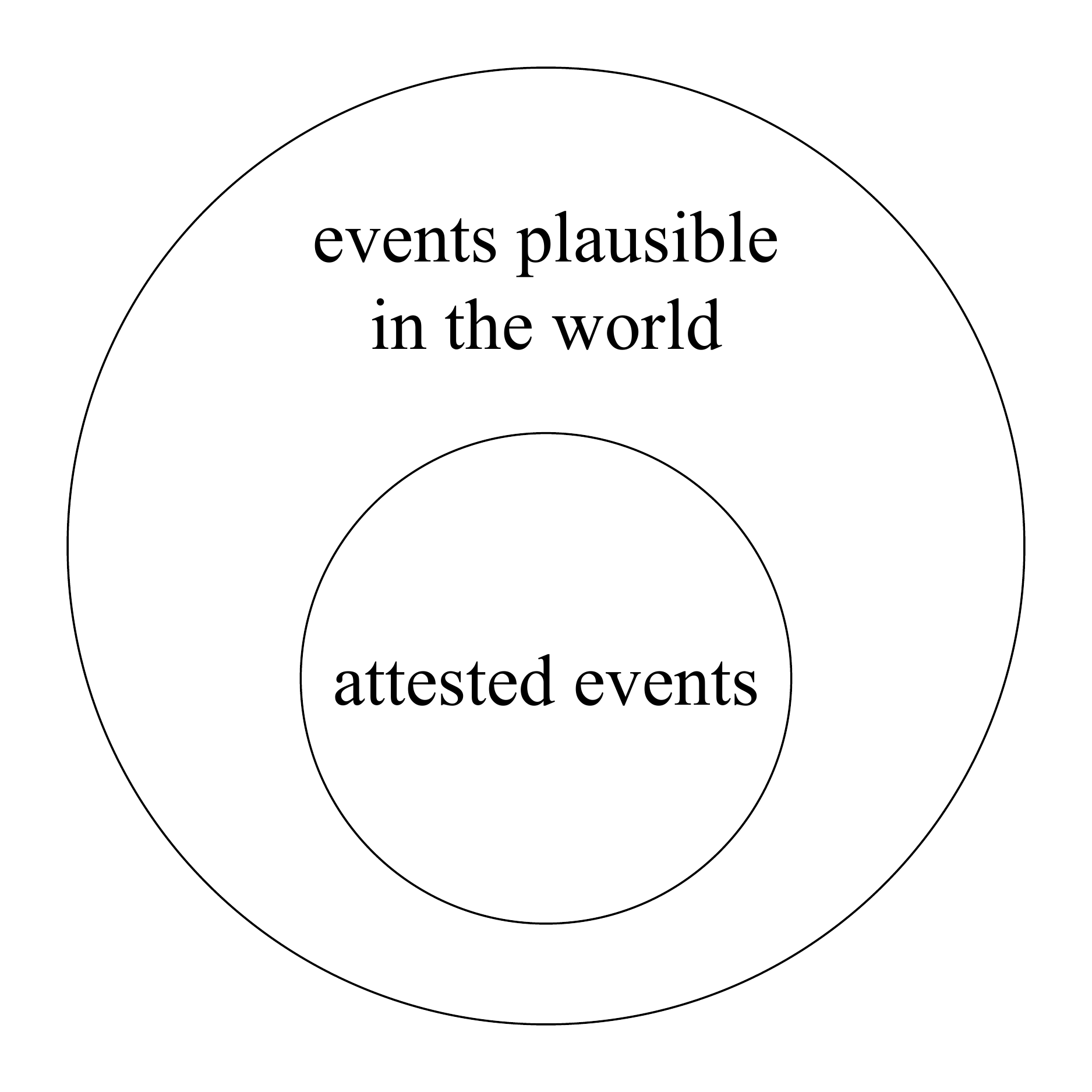}
  \caption{An attested event is necessarily plausible in the world, but not all plausible events are attested. By the world we refer to some possible world under consideration---in this sense plausibility is an epistemic modality.}
  \label{fig:plausibility}
\end{figure}

While plausibility is difficult to define precisely, we adopt the following useful distinctions from the literature:

\begin{itemize}
\setlength\itemsep{0.2em}
    \item Plausibility is a matter of degree \cite{WILKS197553, resnik_selection_1993}. We therefore evaluate models by their ability to estimate the relative plausibility of events.
    \item Plausibility describes non-surprisal conditioned on some context \cite{resnik_selection_1993, Gordon2011SemEval2012T7}. For example, conditioned on the event ``breathing,'' it is less surprising to learn that the agent is ``a dentist'' than ``a thought'' and thus more plausible.
    \item Plausibility is dictated by likelihood of occurrence in the world rather than text \cite{zhang-etal-2017-ordinal,wang-etal-2018-modeling}. This discrepancy is due to reporting bias---the fact that people do not state the obvious \cite{Gordon:2013:RBK:2509558.2509563, shwartz-choi-2020-neural}; e.g., ``a person dying'' is more likely to be attested than ``a person breathing'' (Figure \ref{fig:plausibility}).
\end{itemize}

\citet{wang-etal-2018-modeling} present the problem formulation that we use in this work, and they show that static word embeddings lack the world knowledge needed for modeling plausibility.

The state of the art is to take the conditional probability of co-occurrence as estimated by a distributional model as an approximation of event plausibility \cite{ijcai2020-554}. Our fine-tuned RoBERTa baseline follows this approach.

Similar in spirit to our work, \citet{He2020OnTR} extend this baseline method by creating additional training data using the Probase taxonomy \cite{probase} in order to improve conceptual generalization; specifically, for each training example they swap the event's arguments with its hypernym or hyponym, and they take this new, perturbed example to be an implausible event.

There is also recent work focusing on monotonic inferences in semantic entailment \cite{yanaka-etal-2019-neural,goodwin-etal-2020-probing,geiger-etal-2020-neural}. 
Plausibility contrasts with entailment in that plausibility is not strictly monotonic with respect to hypernymy/hyponymy relations: the plausibility of an entity is not sufficient to infer the plausibility of its hyponyms (i.e., not downward entailing: it is plausible that a person gives birth but not that a man gives birth) nor hypernyms (i.e., not upward entailing: it is plausible that a baby fits inside a shoebox but not that a person does).

Non-monotonic inferences have recently been explored in the context of defeasible reasoning \cite{rudinger-etal-2020-thinking}: inferences that may be strengthened or weakened given additional evidence. The change in plausibility between an event and its abstraction can be formulated as a type of defeasible inference, and our findings may contribute to future work in this area.

\subsection{Selectional Preference}
\label{sec:selpref}

Modeling the plausibility of single events is also studied in the context of \textit{selectional preference}---the semantic preference of a predicate for taking an argument as a particular dependency relation \cite{evens_semantic_1975, resnik_selection_1993, erk-etal-2010-flexible}; e.g., the relative preference of the verb ``breathe'' for the noun ``dentist'' as its nominal subject.

Models of selectional preference are sometimes evaluated by correlation with human judgements \cite{o-seaghdha-2010-latent,zhang-etal-2019-sp}. The primary distinction between such evaluations and those of semantic plausibility, as in our work, is that evaluations of semantic plausibility emphasize the importance of correctly modeling atypical yet plausible events \cite{wang-etal-2018-modeling}.

Closely related to our work are models of selectional preference that use the WordNet hierarchy to generalize co-occurrence probabilities over concepts. These include the work of \citet{resnik_selection_1993}, related WordNet-based models \cite{li-abe-1998-generalizing,clark-weir-2002-class}, and a more recent experiment by \citet{o-seaghdha-korhonen-2012-modelling} to combine distributional models with WordNet. Notably, these methods make a discrete decision as to the right level of abstraction---if the most preferred subject of ``breathe'' is found to be ``person,'' for example, then all hyponyms of ``person'' will be assigned the same selectional preference score.


\subsection{Conceptual Abstraction}
\label{sec:conceptual-abstraction}

Our second proposed method can be thought of as finding the right level of abstraction at which to infer plausibility. This problem has been broadly explored by existing work.

\citet{van-durme-etal-2009-deriving} extract abstracted commonsense knowledge from text using WordNet, obtaining inferences such as ``A [\textsc{person}] can breathe.'' They achieve this by first extracting factoids and then greedily taking the WordNet synset that dominates the occurrences of factoids to be the appropriate abstraction.

\citet{gong-verb-concepts-2016} similarly abstract a verb's arguments into a set of prototypical concepts using Probase and a branch-and-bound algorithm. For a given verb and argument position, their algorithm finds a small set of concepts that has high coverage of all nouns occurring in said position.

Conceptual abstractions are captured to some extent in pre-trained language models' representations \cite{ravichander-etal-2020-systematicity,weir-etal-2020-probing}.





\section{Problem Formulation}

Given a vocabulary of subjects $\mathcal{S}$, verbs $\mathcal{V}$, and objects $\mathcal{O}$, let an event be represented by the s-v-o triple $e \in \mathcal{S} \times \mathcal{V} \times \mathcal{O}$.

We take $g$ to be a ground-truth, total ordering of events expressed by the ordering function $g(e) > g(e')$ iff $e$ is more plausible than $e'$. Our objective is to learn a model $f:\mathcal{S} \times \mathcal{V} \times \mathcal{O} \rightarrow \mathbf{R}$ that is monotonic with respect to $g$, i.e., $g(e)>g(e') \implies f(e)>f(e')$.

This simplification follows from previous work \cite{wang-etal-2018-modeling}, and the plausibility score for a given triple can be considered the relative plausibility of the respective event across all contexts and realizations.

While meaning is sensitive to small linguistic perturbations, we are interested in cases where one event is more plausible than another marginalized over context. Consider that \textit{person-breathe-air} is more plausible than \textit{thought-breathe-car} regardless of the choice of determiners or tense of the verb.

In practice, we would like to learn $f$ without supervised training data, as collecting a sufficiently large dataset of human judgements is prohibitively expensive \cite{zhang-2020-aser}, and supervised models often learn dataset-specific correlations \cite{levy-etal-2015-supervised,gururangan-etal-2018-annotation,poliak-etal-2018-hypothesis,mccoy-etal-2019-right}. Therefore, we train model $f$ with distant supervision and evaluate by correlation with human ratings of plausibility which represent the ground-truth ordering $g$.

\subsection{Lexical Hierarchy}
\label{sec:lexical-hierarchy}

We define $\mathcal{C}$ to be the set of concepts in a lexical hierarchy, in our case \textit{synsets} in WordNet, with some root concept $c^{(1)} \in \mathcal{C}$. The \textit{hypernym chain} of concept $c^{(h)} \in \mathcal{C}$ at depth $h$ in the lexical hierarchy is defined to be the sequence of concepts $\alpha(c^{(h)}) = (c^{(1)}, c^{(2)}, \ldots, c^{(h)})$ where $\forall i, c^{(i)}$ is a direct hypernym of $c^{(i+1)}$. A lexical hierarchy may be an acyclic graph in which case concepts can have multiple hypernyms, and it follows that there may be multiple hypernym chains to the root. In this case, we take the hypernym chain $\alpha(c^{(h)})$ to be the shortest such chain.

\subsection{Consistency Metrics}
\label{sec:consistency-metrics}

Based on our intuition as to how we expect plausibility estimates to be consistent across abstractions in a hypernym chain, we propose two quantitative metrics of \textit{inconsistency}, Concavity Delta (\ccd{}) and Local Extremum Rate (\ler{}). These metrics provide insight into the degree to which a model's estimates are inconsistent.

\subsubsection{Concavity Delta}
\label{sec:concav-delta}

For a given event, as we traverse up the hypernym chain to higher conceptual abstractions, we expect plausibility to increase until we reach some maximally appropriate level of abstraction, and then decrease thereafter. In other words, we expect that consistent estimates will be concave across a sequence of abstractions.

For example, in the sequence of abstractions ``penguin flies'' $\to$ ``bird flies'' $\to$ ``animal flies,'' plausibility first increases and then decreases. Our intuition is that plausibility increases as we approach the most appropriate level of abstraction, then decreases beyond this level.

A concave sequence is defined to be a sequence $(a_1, a_2, a_3, ...)$ where $\forall i, 2a_i > a_{i-1} + a_{i+1}$.

Let $a_{i-1}$, $a_i$, and $a_{i+1}$ be the plausibility estimates for three sequential abstractions of an event. We define the \textit{divergence from concavity} to be
\begin{equation*}
    \delta = 
\begin{cases} 
      \frac{1}{2}(a_{i-1} + a_{i+1}) - a_i & 2a_i < a_{i-1} + a_{i+1} \\
      0 & $otherwise$ \\
   \end{cases}
\end{equation*}
We then define the \textit{Concavity Delta}, \ccd{}, to be the average $\delta$ across all triples of conceptually sequential estimates. Ideally, a model's estimates should have low \ccd{}. A higher \ccd{} reflects the extent to which models violate our intuition.

\subsubsection{Local Extremum Rate}

\ler{} simply describes how often a conceptual abstraction is a local extremum in terms of its plausibility estimate. Most often, the change in plausibility between sequential abstractions is consistently in the same direction. For example, from ``bird flies'' $\to$ ``animal flies'' $\to$ ``organism flies,'' plausibility consistently decreases. The majority of abstractions will not be the most appropriate level of abstraction and therefore not a local extremum.

As in \S\ref{sec:concav-delta}, we consider all triples of conceptually sequential estimates of the form $a_{i-1}$, $a_i$, and $a_{i+1}$. Formally, \ler{} is the number of triples where $a_i > max(a_{i-1}, a_{i+1})$ or $a_i < min(a_{i-1}, a_{i+1})$ divided by the total number of triples.

A high LER signifies that plausibility estimates have few monotonic subsequences across abstractions. Therefore, a more consistent model should have a lower LER. There are, of course, exceptions to our intuition, and this metric is most insightful when it varies greatly between models.

\begin{figure*}
\centering
  \includegraphics[width=16cm,height=7cm]{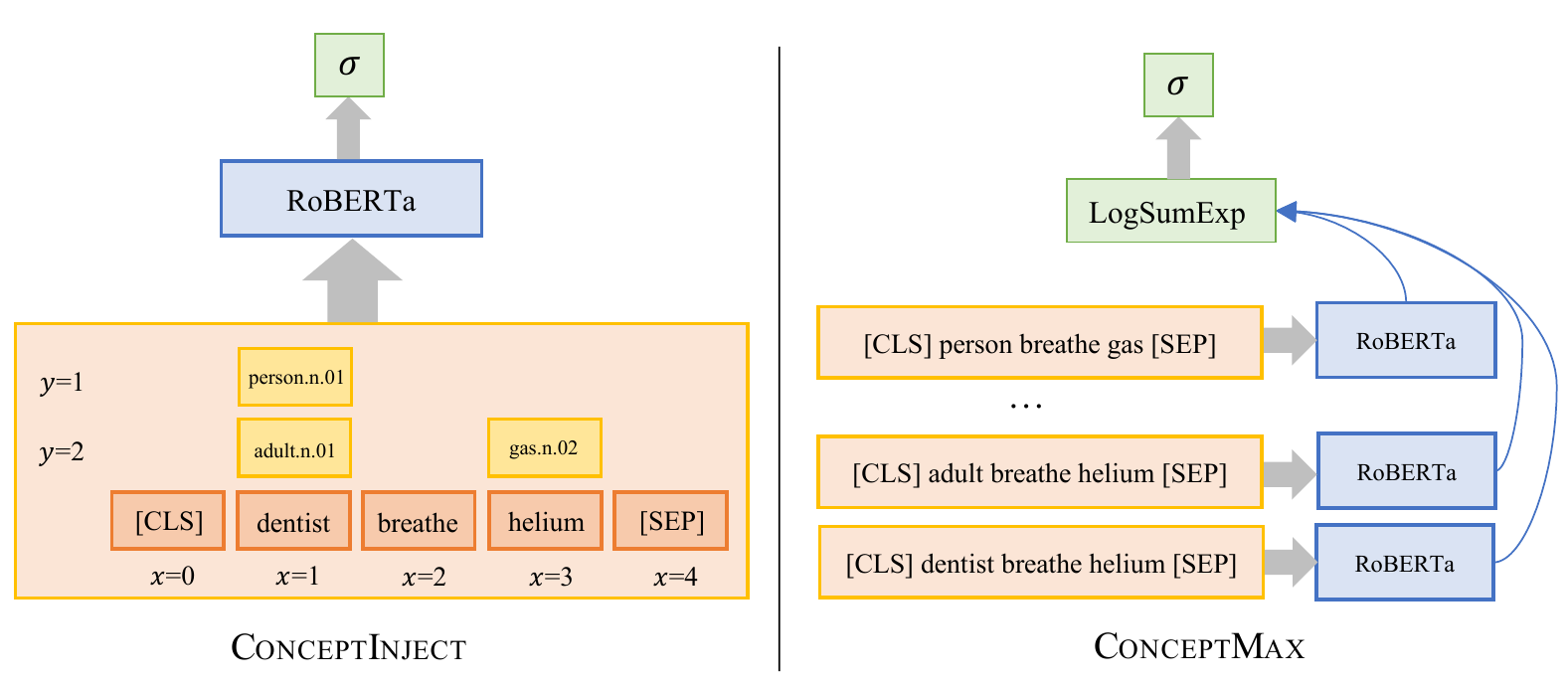}
  \caption{Left: The general formulation of \conceptinject{}; this model takes as input an event and the full hypernym chains of each argument. Right: \conceptmax{} which calculates a plausibility score for each abstraction of an event using RoBERTa, and then takes the ultimate output to be the maximum of these abstractions. $\sigma$ represents an element-wise sigmoid function.}
  \label{fig:concept-inject}
\end{figure*}

\section{Models}

The models that we consider are all of the same general form. They take as input an event and output a relative plausibility score.

\subsection{RoBERTa}
\label{sec:roberta}

Our proposed models are structured on top of a RoBERTa baseline. We use RoBERTa in the standard sequence classification framework. We format an event in the raw form as \texttt{`\small [CLS] subject verb object [SEP]'} where the s-v-o triple is tokenized using a byte pair encoding.\footnote{Technically, RoBERTa's \texttt{[CLS]} and \texttt{[SEP]} tokens are \texttt{<s>} and \texttt{</s>}.} These tokens are used as input to a pre-trained RoBERTa model, and a linear layer is learned during fine-tuning to project the final-layer \texttt{[CLS]} token representation to a single logit which is passed through a sigmoid to obtain the final output, $f(e)$.

We use the HuggingFace Transformers library PyTorch implementation of RoBERTa-base with 16-bit floating point precision \cite{wolf-etal-2020-transformers}.

\subsection{\conceptinject{}}
\label{sec:concept-inject}

\conceptinject{} is an extension of the existing state-of-the-art plausibility models. This model takes as input, in addition to an event, the hypernym chains of the synsets corresponding to each argument in the event. We propose this model to explore how injecting simple awareness of a lexical hierarchy affects estimates.

\conceptinject{} is similar in principle to Onto-LSTM \cite{dasigi-etal-2017-ontology}, which provides the entire hypernym chains of nouns as input to an LSTM for selectional preference, and also similar to K-BERT \cite{Liu_Zhou_Zhao_Wang_Ju_Deng_Wang_2020}, which injects knowledge into BERT during fine-tuning by including relations as additional tokens in the input. K-BERT has demonstrated improved performance over Chinese BERT on several NLP tasks.

The model extends our vanilla RoBERTa baseline (\S\ref{sec:roberta}). We add an additional token embedding to RoBERTa for each synset $c \in \mathcal{C}$. We initialize the embedding of $c$ as the average embedding of the sub-tokens of $c$'s lemma.\footnote{We refer to the name of a synset as the synset's lemma, e.g. the lemma of the synset [dog.n.01] is taken to be ``dog.'' For synsets that correspond to multiple lemmas, we randomly sample one.} We refer to RoBERTa's positional embedding matrix as the $x$-position and randomly initialize a second positional embedding matrix, the $y$-position.

The model input format follows that used for RoBERTa (\S\ref{sec:roberta}), with the critical distinction that we also include the tokens for the hypernyms of the subject and object as additional input.

For the subject $s$, we first disambiguate the synset $c$ of $s$ using BERT-WSD \cite{yap-etal-2020-adapting}. Then for each hypernym $c^{(i)}$ in the hypernym chain $\alpha({c})$, the token of $c^{(i)}$ is included in the model input: this token takes the same $x$-position as the first sub-token of $s$ and takes its $y$-position to be $i$, the depth in the lexical hierarchy.  Finally, the $x$-position, $y$-position, and token embedding are summed for each token to compute its initial representation (Figure~\ref{fig:concept-inject}).

The hypernyms of the object are included by the same procedure. Non-synset tokens have a $y$-position of zero. \conceptinject{} thus sees an event and the full hypernym chains of the arguments when computing a plausibility score.

\subsection{\conceptmax{}}
\label{sec:concept-max}

\conceptmax{} is a simple post-hoc addition to the vanilla RoBERTa model (\S\ref{sec:roberta}). We compute a score for all abstractions of an event $e$ and take the final plausibility $f(e)$ to be a soft maximum of these scores. This method is inspired by that of \citet{resnik_selection_1993} which takes selectional preference to be a hard maximum of some plausibility measure over concepts.

Again, we use BERT-WSD to disambiguate the synset of the subject, $c^{(h)}_s$, and the synset of the object, $c^{(l)}_o$. Using RoBERTa as in \S\ref{sec:roberta}, we then compute a plausibility score for every triple of the form $(c^{(i)}_s,v,c^{(j)}_o)$ where $c^{(i)}_s$ and $c^{(j)}_o$ are hypernyms in the hypernym chains $\alpha(c^{(h)}_s)$ and $\alpha(c^{(l)}_o)$, respectively. Synsets are represented by their lemma when used as input to RoBERTa. Finally, we take the LogSumExp, a soft maximum, of these scores to be the ultimate output of the model (Figure \ref{fig:concept-inject}).

During training, we sample only three of the abstractions $(c^{(i)}_s,v,c^{(j)}_o)$ to reduce time complexity. Thus we only need to compute four total scores instead of $h \times l$. At inference time, we calculate plausibility with a hard maximum over all triples.

\subsection{Additional Baselines}

\paragraph{\robertazs{}} We use MLConjug\footnote{\url{https://pypi.org/project/mlconjug/}} to realize an s-v-o triple in natural language with the determiner ``the'' for both the subject and object, and the verb conjugated in the indicative, third person tense; e.g., \textit{person-breathe-air} $\longrightarrow$ ``The person breathes the air.'' We first mask both the subject and object to compute $P(o|v)$, then mask just the subject to compute $P(s|v,o)$. Finally we calculate $f(e)=P(s,o|v)=P(s|v,o) \cdot P(o|v)$. In the case that a noun corresponds to multiple tokens, we mask all tokens and take the probability of the noun to be the geometric mean of its token probabilities.

\paragraph{\mlp{}} The selectional preference model of \citet{van-de-cruys-2014-neural} initialized with GloVe embeddings \cite{pennington-etal-2014-glove}.

\paragraph{n-gram} A simple baseline that estimates $P(s,o|v)$ by occurrence counts. We use a bigram model as we found trigrams to correlate less with human judgments.
\begin{equation}
    P(s,o|v) \approx \frac{\texttt{Count}(s,v) \cdot \texttt{Count}(v,o)}{\texttt{Count}(v)^2}
\end{equation}


\begin{table}
\centering
\begin{tabular}{cc}
\toprule
$e$ & $e'$ \\
\midrule
\textit{animal-eat-seed} & \textit{animal-eat-area} \\
\addlinespace[3pt]
\textit{passenger-ride-bus} & \textit{bus-ride-bus} \\
\addlinespace[3pt]
\textit{fan-throw-fruit} & \textit{group-throw-number} \\
\addlinespace[3pt]
\textit{woman-seek-shelter} & \textit{line-seek-issue} \\
\bottomrule
\end{tabular}
    \caption{
    Training examples extracted from Wikipedia. Event $e$ is an attested event taken to be more plausible than its random perturbation $e'$.
    }
    \label{table:wikipedia}
\end{table}

\section{Training}
\label{sec:training}

Models are all trained with the same objective to discriminate plausible events from less plausible ones. Given a training set $\mathcal{D}$ of event pairs $(e,e')$ where $e$ is more plausible than $e'$, we minimize the binary cross-entropy loss
\begin{equation}
\label{eq:bce-loss}
    L = - \sum_{(e,e') \in \mathcal{D}} \log(f(e)) + \log (1-f(e'))
\end{equation}

In practice, $\mathcal{D}$ is created without supervised labels. For each ${(e,e') \in \mathcal{D}}$, $e$ is an event attested in a corpus with subject $s$, verb $v$, and object $o$. $e'$ is a random perturbation of $e$ uniformly of the form $(s',v,o)$, $(s,v,o')$, or $(s',v,o')$ where $s'$ and $o'$ are arguments randomly sampled from the training corpus by occurrence frequency. This is a standard pseudo-disambiguation objective. Our training procedure follows recent works that learn plausibility models with self-supervised fine-tuning \cite{kocijan-etal-2019-surprisingly,He2020OnTR,ijcai2020-554}.

For the models that use WordNet, we use a filtered set of synsets: we remove synsets with a depth less than 4, as these are too broad to provide useful generalizations \cite{van-durme-etal-2009-deriving}. We also filter out synsets whose corresponding lemma did not appear in the training corpus.

The WordNet models also require sense disambiguation. We use the raw triple as input to BERT-WSD \cite{yap-etal-2020-adapting} which outputs a probability distribution over senses. We take the argmax to be the correct sense.

We train all models with gradient descent using an Adam optimizer, a learning rate of 2e-5, and a batch size of 128. We train for two epochs over the entire training set of examples with a linear warm-up of the learning rate over the first 10,000 iterations. Fine-tuning RoBERTa takes five hours on a single Nvidia V100 32GB GPU. Fine-tuning \conceptinject{} takes 12 hours and \conceptmax{} 24 hours.

\subsection{Training Data}
\label{sec:training-data}

We use English Wikipedia to construct the self-supervised training data. As a relatively clean, definitional corpus, plausibility models trained on Wikipedia have been shown to correlate with human judgements better than those trained on similarly sized corpora \cite{zhang-etal-2019-sp,porada-etal-2019-gorilla}.

We parse a dump of English Wikipedia using the Stanford neural dependency parser \cite{qi2018universal}. For each sentence with a direct object, no indirect object, and noun arguments (that are not proper nouns), we extract a training example $(s,v,o)$: we take $s$ and $o$ to be the lemma of the head of the respective relations (\texttt{nsubj} and \texttt{obj}), and $v$ to be the lemma of the head of the root verb. This results in some false positives such as the sentence ``The woman eats a hot dog.'' being extracted to the triple \textit{woman-eat-dog} (Table \ref{table:wikipedia}).

We filter out triples that occur less than once and those where a word occurred less than 1,000 times in its respective position. We do not extract the same triple more than 1,000 times so as not to over-sample common events. In total, we extract 3,298,396 triples (representing 538,877 unique events).

\section{Predicting Human Plausibility Judgements}
\label{sec:evaluation}

We evaluate models by their correlation with human plausibility judgements. Each dataset consists of events that have been manually labelled to be plausible or implausible (Table \ref{table:human-plaus-judge}). We use AUC (area under the receiver-operating-characteristic curve) as an evaluation metric which intuitively reflects the ability of a model to discriminate a plausible event from an implausible one.

These datasets contain plausible events that are both typical and atypical. While a distributional model should be able to discriminate typical events given that they frequently occur in text, discriminating atypical events (such as \textit{dentist-breathe-helium}) is more difficult.

\begin{table}
\centering
\begin{tabular}{ccc}
\toprule
Topic & Question & Answer \\
\midrule
cat & Does it lay eggs? & never \\
\addlinespace[3pt]
carrot & Can you eat it? & always \\
\addlinespace[3pt]
cocoon & Can it change shape? & sometimes \\
\addlinespace[3pt]
clock & Can I touch it? & always \\
\bottomrule
\end{tabular}
    \caption{
    Example triples from the 20 Questions commonsense dataset. These are those specific examples that contain a simple question with a single s-v-o triple and no modifiers.
    }
    \label{table:twentyquestions}
\end{table}

\begin{table}
\centering
\begin{tabular}{c@{\hskip 2em}cc}
\toprule
\multirow{4}{*}{\pep{}} &  {\it chef-bake-cookie} & \raisebox{0pt}{\small \Checkmark} \\
& {\it dog-close-door} & \raisebox{0pt}{\small \Checkmark} \\
& {\it fish-throw-elephant} & \raisebox{-2pt}{\small \XSolidBrush} \\
& {\it marker-fuse-house} & \raisebox{-2pt}{\small \XSolidBrush} \\
\addlinespace[2pt]
\hline
\addlinespace[2pt]
\multirow{4}{*}{20Q} & {\it whale-breathe-air} & \raisebox{0pt}{\small \Checkmark} \\
& {\it wolf-wear-collar} & \raisebox{0pt}{\small \Checkmark} \\
& {\it cat-hatch-egg} & \raisebox{-2pt}{\small \XSolidBrush} \\
& {\it armrest-breathe-air} & \raisebox{-2pt}{\small \XSolidBrush} \\
\bottomrule
\end{tabular}
    \caption{Representative examples taken from the validation splits of the two plausibility evaluation datasets, \pep{} and 20Q. For simplicity, we present human judgments as plausible ({\small \Checkmark}) or implausible (\raisebox{-1pt}{\small \XSolidBrush}). Details are provided in \S\ref{sec:evaluation}. 
    }
    \label{table:human-plaus-judge}
\end{table}

\subsection{\pep{}}

\textsc{Pep-3k}, the crowdsourced \textbf{P}hysical \textbf{E}vent \textbf{P}lausbility ratings of \citet{wang-etal-2018-modeling}, consists of 3,062 events rated as physically plausible or implausible by five crowdsourced workers. Annotators were instructed to ignore possible metaphorical meanings of an event. We divide the dataset equally into a validation and test set following the split of \citet{porada-etal-2019-gorilla}.

To evaluate on this dataset, we make the assumption that all events labeled physically plausible are necessarily more plausible than all those labeled physically implausible.

\subsection{20Q}

The 20 Questions commonsense dataset\footnote{\url{https://github.com/allenai/twentyquestions}} is a collection of 20 Questions style games played by crowdsourced workers. We format this dataset as plausibility judgments of s-v-o triples similar to \pep{}.

In the game 20 Questions, there are two players---one who knows a given topic, and the other who is trying to guess this topic by asking questions that have a discrete answer. The dataset thus consists of triples of topics, questions, and answers where the answer is one of: always, usually, sometimes, rarely, or never (Table \ref{table:twentyquestions}).

We parse the dataset using the Stanford neural dependency parser \cite{qi2018universal}. We then extract questions that contain a simple s-v-o triple with no modifiers where either the subject or object is a third person singular pronoun. We replace this pronoun with the topic, and otherwise replace any occurrence of a personal pronoun with the word ``person.'' We filter out examples where only two of three annotators labelled the likelihood as never. Finally, we take events labelled ``never'' to be less plausible than all other events. This process results in 5,096 examples equally divided between plausible and implausible. We split examples into equal sized validation and test sets.

\begin{table}
\centering
\begin{tabular}{@{\extracolsep{0mm}}
                @{}l
                @{\hspace{5mm}} c
                @{\hspace{5mm}} c
                @{\hspace{7mm}} c @{}}
\toprule   
Model & \pep{} & 20Q & Avg. \\
\midrule
n-gram & .51 & .52 & .52 \\
\mlp{} & .55 & .52 & .53 \\
\robertazs{} & .56 & .57 & .56 \\
RoBERTa & .64 & .67 & .66 \\
\midrule
\conceptinject{} & .64 & .66 & .65 \\
\conceptmax{} & \textbf{.67} & \textbf{.74} & \textbf{.70} \\
\bottomrule
\end{tabular}
\caption{Test set results for predicting human plausibility judgements. Performance is evaluated with AUC with respect to the ground-truth, manually labeled plausibility ratings.}
\label{table:results}
\end{table}

\subsection{Quantitative Results}

Despite making a discrete decision about the right level of abstraction, \conceptmax{} has higher AUC on both evaluation sets as compared to \conceptinject{} and the vanilla RoBERTa baseline (Table~\ref{table:results}). The fact that the \conceptmax{} model aligns with human judgments more than the baselines supports the hypothesis that conceptual consistency improves plausibility estimates.

\conceptinject{} performs similarly to the RoBERTa baseline even though this model is aware of the WordNet hierarchy. We hypothesize that the self-supervised learning signal does not incentivize use of this hierarchical information in a way that would increase correlation with plausibility judgements. We do find that \conceptinject{} attends to the hypernym chain, however, by qualitatively observing the self-attention weights.

All fine-tuned RoBERTa models correlate better with plausibility judgements than the \robertazs{} baseline, and the n-gram baseline performs close to random---this is perhaps to be expected, as very few of the evaluation triples occur in our Wikipedia training data.

\subsection{Qualitative Analysis}

To better understand the performance of these models, we manually inspect 100 examples from each dataset. We find that RoBERTa rarely assigns a high score to a nonsensical event (although this does occur in five cases, such as \textit{turtle-climb-wind} and \textit{person-throw-library}). RoBERTa also rarely assigns a low score to a seemingly typical event, although this is somewhat more common (in cases such as \textit{kid-use-handbag} and \textit{basket-hold-clothes}, for example). This finding confirms our expectation that discerning the typical and nonsensical should be relatively easy for a distributional model.

Examples not at the extremes of plausibility are harder to categorize; however, one common failure seems to be when the plausibility of an event hinges on the relative size of the subject and object, such as in the case of \textit{dog-throw-whale}. This finding is similar to the limitations of static word embeddings observed by \citet{wang-etal-2018-modeling}.

\begin{figure*}%
    \centering
    \subfloat[\centering RoBERTa]{{\includegraphics[width=5cm]{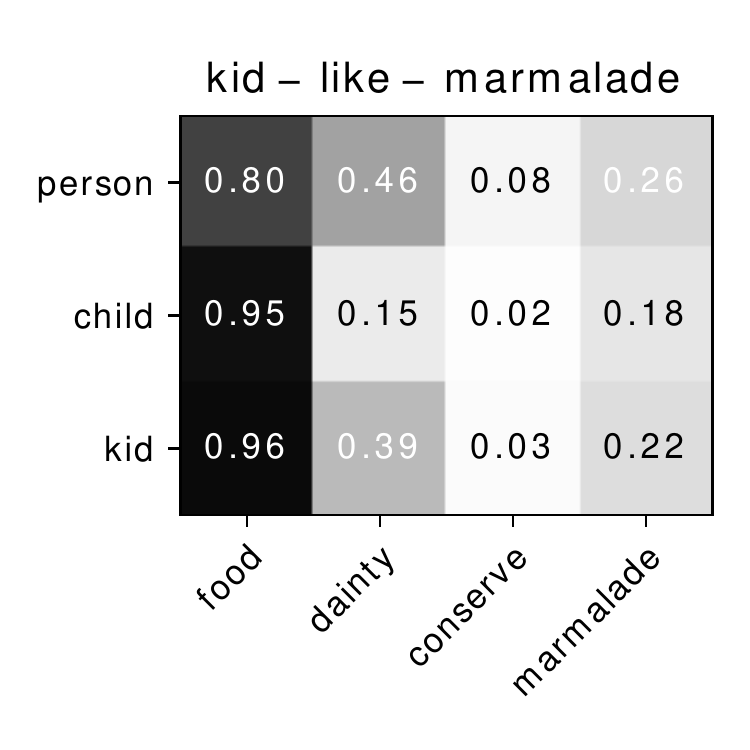} }}%
    \subfloat[\centering \conceptinject{}]{{\includegraphics[width=5cm]{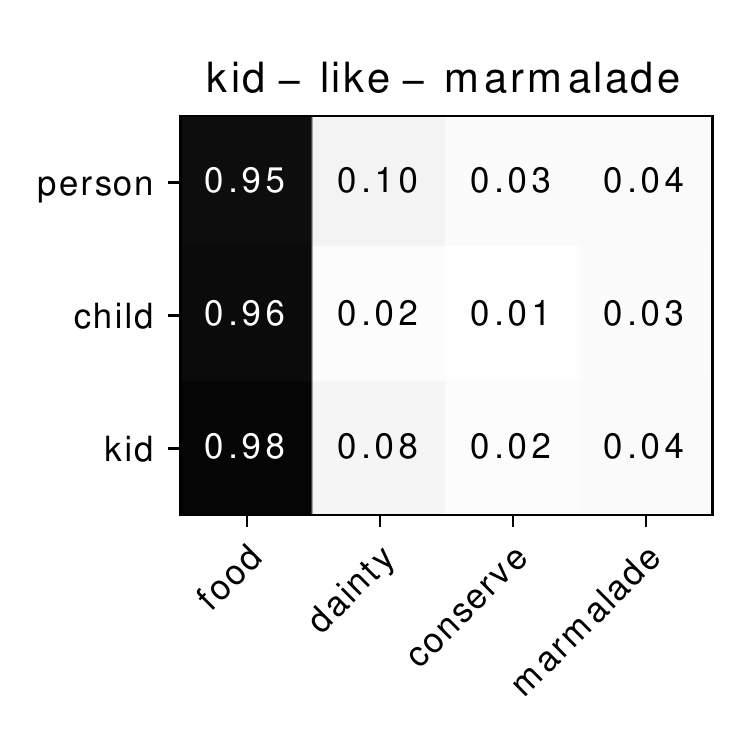}}}
    \subfloat[\centering \conceptmax{}]{{\includegraphics[width=5cm]{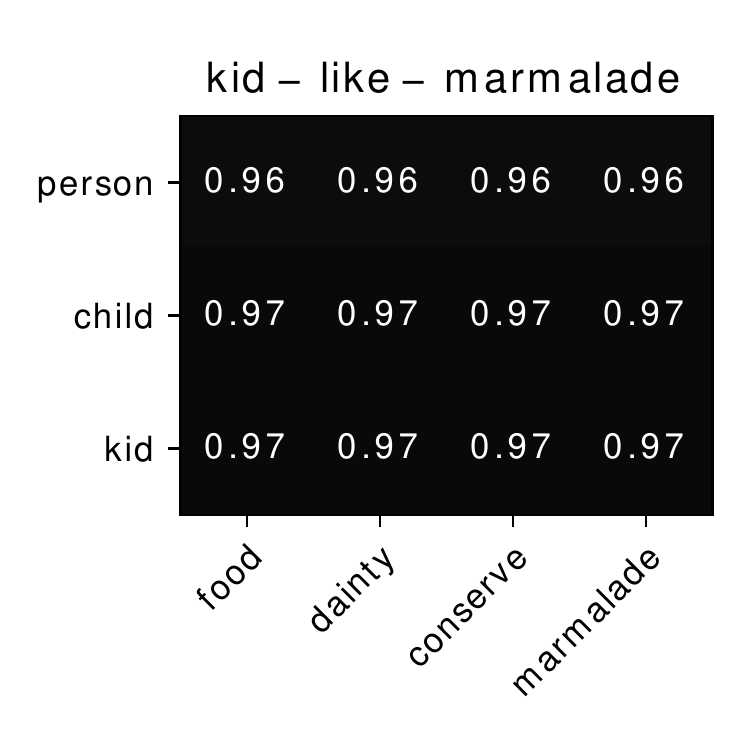}}}%
    \caption{Outputs across conceptual abstractions for the event \textit{kid-like-marmalade} from the 20Q dataset. This event is taken to be relatively plausible as the ground-truth label was ``usually.''}%
    \label{fig:roberta-results}%
\end{figure*}

\begin{table}
\centering
\begin{tabular}{@{\extracolsep{0mm}}
                @{}l
                @{\hspace{5mm}} c
                @{\hspace{\lengtha}} c
                @{\hspace{\lengthb}} c
                @{\hspace{\lengtha}} c @{}}
\toprule   
{} & \multicolumn{2}{c@{\hspace{\lengthc}}}{\pep{}} & \multicolumn{2}{c}{20Q} \\
 \cmidrule(r{\lengthb}){2-3}
 \cmidrule(){4-5}
Model & \ccd{} & \ler{} & \ccd{} & \ler{} \\ 
\midrule
n-gram & .06 & .50 & .07 & .50 \\
\mlp{} & .03 & .61 & .03 & .49 \\
\robertazs{} & .13 & .70 & .12 & .65 \\
RoBERTa & .09 & .52 & .08 & .51 \\
\midrule
\conceptinject{} & .08 & .52 & .07 & .51 \\
\conceptmax{} & .02 & .00 & .02 & .00 \\
\bottomrule
\end{tabular}
\caption{Evaluation of inconsistency. \ccd{} describes the degree to which sequences of estimates across a hypernym chain diverge from a concave sequence. \ler{} describes how often conceptual abstractions are local extrema with respect to plausibility.}
\label{table:inconsistency}
\end{table}

\section{Consistency Evaluation}
\label{sec:consistency}

For every event $e$ in the evaluation sets of human plausibility judgments (\S\ref{sec:evaluation}), we disambiguate $e$ using BERT-WSD and then calculate models' estimates for the plausibility of every possible abstraction of $e$ (Figure \ref{fig:roberta-results}). Based on these estimates, we can analyze the consistency of each model across abstractions.

\subsection{Quantitative Results}

We use our proposed metrics of consistency (\S\ref{sec:consistency-metrics}) to evaluate the extent to which models' estimates are consistent across a hypernym chain (Table \ref{table:inconsistency}).

\robertazs{}, which correlates with plausibility the least of the RoBERTa models, has by far the highest inconsistency.

The fine-tuned RoBERTa and \conceptinject{} estimates are also largely inconsistent by our metrics. For these models, half of all estimates are a local extrema in the lexical hierarchy. As shown in Figure \ref{fig:roberta-results}, the space of plausibility estimates is rigid for these models, and most estimates are a local extremum with respect to the plausibility of the subject or object of the event.

\conceptmax{} is almost entirely consistent by these metrics, which is to be expected as this model makes use of the same WordNet hierarchy that we are using for evaluation. We also evaluated consistency using the longest rather than the shortest hypernym chain in WordNet, but did not find a significant change in results. This is likely because for the consistency evaluation we are using the hypernym chains that have been filtered as described in \S\ref{sec:lexical-hierarchy}.

\subsection{Qualitative Observations}

We qualitatively evaluate the consistency of models by observing the matrix of plausibility estimates for all abstractions as show in Figure \ref{fig:roberta-results}.

In agreement with our quantitative metrics, we observe that RoBERTa estimates are often inconsistent in that they vary greatly between two abstractions that have similar plausibility. Surprisingly, however, it is also often the case that RoBERTa estimates are similar or identical between abstractions. In some cases, this may be the result of the model being invariant to the subject or object of a given event.

We also observe the individual examples with the highest \ccd{}. In these cases, it does appear that the variance of model estimates is unreasonable. In contrast, \ler{} is sometimes high for an example where the estimates are reasonably consistent. This is a limitation of the \ler{} metric not taking into account the degree of change between estimates.

Finally, we observe that the BERT-WSD sense is often different from what an annotator primed to rate plausibility would assume. For example, in the case of \textit{dog-cook-turkey}, BERT-WSD takes dog to be a hyponym of person. While this is reasonable in context, it results in a different plausibility than that annotated.

\section{Conclusion}

While the state of the art in modeling plausibility has improved in recent years, models still fall short of human ability. We show that model estimates are inconsistent with respect to a lexical hierarchy: they correlate less with human judgments as compared to model estimates that are forced to be consistent, and they do not satisfy our intuitively defined quantitative measures of consistency.

In addition, we show that simply injecting lexical knowledge into a model is not sufficient to correct this limitation. Conceptual consistency appears to require a more discrete, hierarchical bias.

Interesting questions for future work are: 1) can we design a \textit{non-monotonic}, consistent model of plausibility that better correlates with human judgements? 2) Can we induce a hierarchy of abstractions rather than using a manually created lexical hierarchy?

\section*{Acknowledgments}
We would like to thank Ali Emami and Abhilasha Ravichander for useful discussions and comments, as well as the anonymous reviewers for their insightful feedback.
This work is supported by funding from Microsoft Research and resources from Compute Canada. The last author is supported by the Canada CIFAR AI Chair program.

\bibliography{anthology,custom}
\bibliographystyle{acl_natbib}

\end{document}